\documentclass[letterpaper, 10 pt, conference]{ieeeconf}  % Comment this line out if you need a4paper

\IEEEoverridecommandlockouts                              % This command is only needed if 
\usepackage{graphicx}      % include this line if your document contains figures
\usepackage{amsmath,amssymb,bm,mathtools}
\usepackage{siunitx}
\usepackage{xcolor}
\usepackage{fp, tikz, pgfplots}
\usepackage{algorithmic}
\usepackage{algorithm}
\usepackage[capitalize]{cleveref}
\usepackage{booktabs} % for professional tables
\usepackage{tabularx}
\usetikzlibrary{arrows,shapes,backgrounds,patterns,fadings,matrix,arrows,calc,
	intersections,decorations.markings,
	positioning,external,arrows.meta}
\usepgfplotslibrary{fillbetween}
\definecolor{nicegreen}{RGB}{0,200,0} 

\newtheorem{assumption}{Assumption}

\newtheorem{remark}{Remark}

\pgfplotsset{width=5\columnwidth /5, compat = 1.13, 
	height = 2\columnwidth /5, grid= major, 
	legend cell align = left, ticklabel style = {font=\scriptsize},
	every axis label/.append style={font=\small},
	legend style = {font=\tiny},title style={yshift=-7pt, font = \small} }

\title{\LARGE \bf
Personalized Rehabilitation Robotics based on Online Learning Control%
}

\author{Samuel Tesfazgi$^{1*}$, Armin Lederer$^{1*}$, Johannes F. Kunz, Alejandro J. Ord\'o\~nez Conejo, Sandra Hirche$^{1}$% <-this % stops a space 
\thanks{$^{1}$Samuel Tesfazgi, Armin Lederer and Sandra Hirche are with the Chair of Information-oriented Control (ITR), Department of Electrical and Computer Engineering, Technical University of Munich, Germany
        {\tt\small \{armin.lederer,samuel.tesfazgi,hirche\}@tum.de}}%
 \thanks{$^{*}$These authors contributed equally.}
}

\begin{document}

\maketitle
\thispagestyle{empty}
\pagestyle{empty}

%%%%%%%%%%%%%%%%%%%%%%%%%%%%%%%%%%%%%%%%%%%%%%%%%%%%%%%%%%%%%%%%%%%%%%%%%%%%%%%%
\begin{abstract}

The use of rehabilitation robotics in clinical applications gains increasing importance, due to therapeutic benefits and the ability to alleviate labor-intensive works. However, their practical utility is dependent on the deployment of appropriate control algorithms, which adapt the level of task-assistance according to each individual patient's need. Generally, the required personalization is achieved through manual tuning by clinicians, which is cumbersome and error-prone. In this work we propose a novel online learning control architecture, which is able to personalize the control force at run time to each individual user. To this end, we deploy Gaussian process-based online learning with previously unseen prediction and update rates. Finally, we evaluate our method in an experimental user study, where the learning controller is shown to provide personalized control, while also obtaining safe interaction forces.

\end{abstract}

%%%%%%%%%%%%%%%%%%%%%%%%%%%%%%%%%%%%%%%%%%%%%%%%%%%%%%%%%%%%%%%%%%%%%%%%%%%%%%%%
\section{INTRODUCTION}\label{sec:intro}

\setlength{\textfloatsep}{8pt}
\setlength{\floatsep}{7pt}
\setlength{\abovedisplayskip}{5pt}
\setlength{\belowdisplayskip}{6pt}

In recent years neurological disorders have become more dominant with an estimate of over 16 million people suffering a first stroke each year \cite{Feigin2014}. Therefore, an urgent need for rehabilitation treatments arises to ensure the quality of living for such patients. In particular high-intensity and repetition training  has been shown to produce the most promising recovery results \cite{Ringleb2008}. Due to these requirements, effective rehabilitation is labor intensive and both patients and healthcare professionals can benefit greatly from robotic-assisted rehabilitation strategies \cite{Krebs2003}. 
However, the control of these devices presents certain challenges, which can limit their applicability in practice. Different factors, such as level of assistance, patient engagement and task success, have to be considered when designing the controller. 
These requirements are particularly difficult to fulfill, due to the uncertain interaction dynamics between human and robot. This problem is exacerbated by the variety of patient behaviors and needs, which require a high degree of personalization. \looseness=-1

A control approach that is widely used in the literature is impedance control, which has previously been shown to be applicable for robot-based arm rehabilitation \cite{Yang2006}. 
Impedance control is particular popular for human-robot interaction, since it provides compliant behavior for appropriately chosen parameters, therefore, ensuring limited and safe interaction forces. In \cite{Zhang2015}, a position-dependent stiffness is used to assist rehabilitation tasks, where the stability of the human-robot interaction is guaranteed using system passivity. Differently, in \cite{Li2018}, a desired impedance model is defined, which is subsequently achieved through an iterative learning scheme.
However, despite these application examples of impedance control for robot-based neurorehabilitation, they generally hinge 
on properly chosen impedance parameters, which first have to be tuned by the manufacturer a-priori and are then readjusted by clinicians \cite{Zhang2015}. This procedure is time-consuming and error prone, due to the variety of target personas. Additionally, the tuning needs to be performed cautiously in order to retain the compliant behavior needed for safe interaction. To address these issues other works have developed control architectures that specifically take the complete dynamics into consideration, e.g., via feed-forward control and disturbance observers \cite{Just18}. While this approach decreases interaction forces, it requires measurements of the patient-robot interaction wrench at each interaction point, such that undesirable force-torque sensors become necessary, which increase system costs and may even lead to stability concerns \cite{Chae87}. \looseness=-1

Alternatively, learning-based controllers can be deployed to fulfill the personalization requirements. To deal with the unknown interaction model and to adapt the patient-robot interaction, He et al. \cite{He2015} propose a neural network control approach for rehabilitation robotics. Differently, the authors in \cite{Medina2019} focus on obtaining human motor control models during physical interaction by deploying Gaussian processes (GPs) to learn the human arm impedance.  
Despite these approaches showing promising results, they cannot account for behavior changes of the patient, i.e., due to fatigue, since the learning is performed offline. However, facilitating this adaptation can be highly beneficial for the personalization of control strategies in rehabilitation robotics \cite{Beckerle2017}.

In this work we present a novel learning-based control architecture that facilitates highly personalized assistance in a human-robot collaboration task, while requiring no parameter tuning. To the best of our knowledge, this is the first time that a GP is used directly in the control loop to generate assistive forces during physical human-robot interaction. Additionally, we evaluate our learning-based control architecture experimentally, where update and prediction rates with online-generated data are achieved that are orders of magnitude higher than previous GP approaches. The remainder of the paper is structured as follows: First the problem statement is introduced in \cref{sec:problem}, while the online learning approach is presented in \cref{sec:online GPs}. Subsequently, in \cref{sec:theory}, the online learning control architecture is proposed. Finally, the evaluation of our method follows in \cref{sec:experiment}. \looseness=-1

\section{PROBLEM FORMULATION}\label{sec:problem}

We model the physical-human robot interaction in rehabilitation robotics using
Euler-Lagrange systems of the form%\footnote{Notation: }
% \pagebreak
\begin{align}\label{eq:sys}
    \bm{H}(\bm{q})\ddot{\bm{q}}+\bm{C}(\bm{q},\dot{\bm{q}})\dot{\bm{q}}+\bm{g}(\bm{q})+\bm{f}_i(\bm{x})=\bm{u},
\end{align}
where $\bm{x}\!=\!\begin{bmatrix}\bm{q}^T&\dot{\bm{q}}^T&\ddot{\bm{q}}^T\ t  \end{bmatrix}^T$ is the concatenation of  joint angles $\bm{q}\in\mathbb{R}^d$, angular velocities $\dot{\bm{q}}\in\mathbb{R}^d$, angular accelerations $\ddot{\bm{q}}\in\mathbb{R}^d$, and the time $t\in\mathbb{R}_{0,+}$. The matrix 
$\bm{H}:\mathbb{R}^d\rightarrow \mathbb{R}^{d\times d}$ denotes the symmetric and positive definite generalized
inertia of the robotic system, $\bm{C}: \mathbb{R}^d\times\mathbb{R}^d\rightarrow\mathbb{R}^{d\times d}$ is the generalized Coriolis matrix, $\bm{g}: \mathbb{R}^d\rightarrow\mathbb{R}^d$ are the torques resulting from gravitation, $\bm{f}_i:\mathbb{R}^{3d+1}\rightarrow\mathbb{R}^d$ describes the interaction torques generated by the $i$-th patient, and $\bm{u}\in\mathbb{R}^d$ are torques applied to the system as control input. 
\begin{remark}
We account for the intra-patient variation of interaction dynamics, e.g., caused by unobserved internal dynamics in the patient such as fatigue, by considering time-dependent functions $\bm{f}_i(\cdot)$. Hence, all behavioral changes of the human are 
modeled through time-dependency.
\end{remark}

In order to reflect the practical 
availability of models, we make the following assumption.
\begin{assumption}\label{ass:knowledgde}
All parameters of the robotic system are known, hence, $\bm{H}(\cdot)$, $\bm{C}(\cdot,\cdot)$ and $\bm{g}(\cdot)$ are available. In contrast, the individual dynamics of patients $\bm{f}_i(\cdot)$ are unknown.
\end{assumption}

This assumption reflects the fact that accurate models of robotic systems can typically be obtained using
classical identification procedures \cite{Hollerbach2008}, which can be directly applied in control design. 
In contrast, the identification of models of human motor dynamics in physical interaction is a challenging
problem and often limited to simple scenarios~\cite{Medina2019}, making them generally inapplicable in rehabilitation 
robotics. 

In order to overcome these limitations of conventional modeling techniques, we employ a non-parametric, 
data-driven approach for learning the human induced dynamics~$\bm{f}_i(\cdot)$. For the
inference of an individual model for each patient, we consider access to the following measurements.
\begin{assumption}\label{ass:measure}
The control input $\bm{u}$, the joint angles $\bm{q}$, and the angular velocities $\dot{\bm{q}}$ 
and accelerations $\ddot{\bm{q}}$ of each patient can be observed for learning a personalized model.\looseness=-1
\end{assumption}

As the control input $\bm{u}$ is determined by the employed control law, it can be directly observed. 
The joint angles~$\bm{q}$ are usually measured, such that the angular velocities~$\dot{\bm{q}}$ and 
accelerations~$\ddot{\bm{q}}$ can be obtained through numerical differentiation. It is important to note
that we do not require force torque sensors to determine interaction forces, which often suffer from 
high measurement noise and are expensive.

The task of each patient is the execution of a rehabilitation exercise described by a reference 
trajectory $\bm{q}_{\mathrm{ref}}$ for system~\eqref{eq:sys}, for which we require the following property.
\begin{assumption}\label{ass:ref}
The bounded reference trajectory $\bm{q}_{\mathrm{ref}}$ is twice continuously differentiable with bounded
derivatives.
\end{assumption}

Since abrupt movements must be avoided in physical human-robot interaction, in particular when dealing
with impaired patients, the required smoothness of the reference is a natural
assumption. Moreover, the assumed boundedness of reference trajectories 
directly follows from the compact work spaces robots and humans operate in, such that \cref{ass:ref}
is not restrictive in practice. 

In order to execute the task by tracking the reference trajectory with the human-robot system \eqref{eq:sys},
a control law needs to be defined to determine the control inputs $\bm{u}$. On the one hand, the applicability in 
real-world rehabilitation robotic scenarios requires this control law to ensure safe interaction
forces by avoiding excessively high control inputs $\bm{u}$. On the other hand, the successful 
execution of the rehabilitation exercise requires a satisfactory tracking accuracy. These requirements
generally pose conflicting goals, which must be traded-off in a control gain tuning phase. 
However, this tuning process cannot be performed for each patient individually, 
such that control gains must be employed, which are expected to perform well on a wide range of patients, 
but can yield rather poor performance for some of them. In order to overcome this issue, 
we consider the problem of designing a control law which adapts online 
to the observed behavior of individual patients using non-parameteric machine learning, such that a
highly personalized treatment can be realized. 

%%%%%%%%%%%%%%%%%%%%%%%%%%%%%%%%%%%%%%%%%%%%%%%%%%%%%%%%%%%%%%%%%%%%%%%%%%%%%%%%%%%%%%%%%%%%%%%%%%%%%%%%%%%%%%%%%%
%%%%%%%%%%%%%%%%%%%%%%%%%%%%%%%%%%%%%%%%%%%%%%%%%%%%%%%%%%%%%%%%%%%%%%%%%%%%%%%%%%%%%%%%%%%%%%%%%%%%%%%%%%%%%%%%%%
\section{GAUSSIAN PROCESS-BASED ONLINE LEARNING}
\label{sec:online GPs}

In order to develop control laws achieving the posed design goals, we employ Gaussian process-based 
machine learning to infer a model of the human motor behavior. The foundations of Gaussian process 
regression are introduced in \cref{subsubsec:GP}, before a Gaussian process based online learning
algorithm  relying on the aggregation of
local Gaussian process models is presented in \cref{subsec:log-GP}.

\subsection{Gaussian Process Regression}\label{subsubsec:GP}

Gaussian process (GP) regression bases on the assumption that any finite number of evaluations
$\{f(\bm{x}^{(n)})\}_{n=1}^N$ of a scalar function $f:\mathbb{R}^{\rho}\rightarrow\mathbb{R}$, $\rho\in\mathbb{N}$, 
at inputs $\bm{x}^{(n)}\in\mathbb{R}^{\rho}$ follows a joint Gaussian distribution \cite{Rasmussen2006}. 
This distribution, denoted as $\mathcal{GP}(m(\bm{x}),k(\bm{x},\bm{x}'))$, 
is defined in terms of a prior mean function $m:\mathbb{R}^{\rho}\rightarrow\mathbb{R}$, which can be 
used to incorporate prior knowledge such as parametric models, and a covariance 
function $k:\mathbb{R}^{\rho}\times\mathbb{R}^{\rho}\rightarrow\mathbb{R}$, capturing more 
abstract information such as differentiability or periodicity of $f(\cdot)$. If no specific knowledge about 
the unknown function $f(\cdot)$ is available, $m(\cdot)$ is commonly set to $0$, which we also assume in the 
remainder of the work. The most frequently used covariance function is the squared exponential kernel 
\begin{align}\label{eq:SE}
	k(\bm{x},\bm{x}')=\sigma_f^2\mathrm{exp}\left(-\sum\limits_{i=1}^{\rho}\frac{(x_i-x_i')^2}{2l_i^2}\right),
\end{align}
whose shape depends on the signal standard deviation $\sigma_f\in\mathbb{R}_+$ and the length scales $l_i\in\mathbb{R}_+$.
The signal standard deviation $\sigma_f$ and the kernel length scales $l_i$, $i=1,\ldots,\rho$, form the hyperparameters $\bm{\theta}=\begin{bmatrix}
\sigma_f& l_1 & \ldots & l_{\rho} & \sigma_{\mathrm{on}}\end{bmatrix}^T$ together 
with an assumed target noise standard deviation $\sigma_{\mathrm{on}}\in\mathbb{R}_+$. The 
hyperparameters $\bm{\theta}$ are commonly obtained by maximizing the log-likelihood%\looseness=-1
% \pagebreak
\begin{align}
	\log p(\bm{y}|\bm{X},\bm{\theta}) \!=\!& \!-\!\frac{1}{2}\bm{y}^T(\bm{K}\!+\!\sigma_{\mathrm{on}}^2\bm{I}_N)^{-1}\bm{y}\nonumber\\
	&\!-\! \frac{1}{2}\log(\det(\mathbf{K}\!+\!\sigma_{\mathrm{on}}^2\bm{I}_N)) \!-\! \frac{n}{2}\log(2\pi),
\end{align}
where we define the elements of the kernel matrix $\bm{K}\!\in\!\mathbb{R}^{N\times N}$ as 
$K_{i,j}\!=\!k(\bm{x}^{(i)},\bm{x}^{(j)})$, and concatenate the training inputs and targets into 
$\bm{X}\!=\!\begin{bmatrix} \bm{x}^{(1)}&\ldots&\bm{x}^{(N)} \end{bmatrix}$ and $\bm{y}\!=\!\begin{bmatrix} y^{(1)}&\ldots&y^{(N)} \end{bmatrix}^T$, respectively. Although this maximization involves a 
non-convex optimization problem, it is typically solved using gradient based optimization methods~\cite{Rasmussen2006}.%\looseness=-1

After the hyperparameters have been optimized, the posterior distribution can be exactly  
calculated under the assumption of training targets perturbed by zero mean Gaussian noise with 
variance $\sigma_{\mathrm{on}}^2$.  This posterior is again Gaussian with mean and variance 
\begin{align}\label{eq:GPmean}
	\mu(\bm{x})&= \bm{y}^T\left(\bm{K}+\sigma_{\mathrm{on}}^2\bm{I}_N \right)^{-1}\bm{k}(\bm{x})\\
	\sigma^2(\bm{x})&=k(\bm{x},\bm{x})-\bm{k}^T(\bm{x})\left(\bm{K}+\sigma_{\mathrm{on}}^2\bm{I}_N \right)^{-1}\bm{k}(\bm{x}),
	\label{eq:GPvariance}
\end{align}
where we define the elements of the kernel vector $\bm{k}(\bm{x})\in\mathbb{R}^{N}$ 
as $k_i(\bm{x})=k(\bm{x}^{(i)},\bm{x})$.

\subsection{Locally Growing Random Trees of Gaussian Processes}\label{subsec:log-GP}

When applying Gaussian process regression in a control application, data becomes available
sequentially as time proceeds. While Gaussian process regression can be straightforwardly 
applied to such data streams using rank one updates in principle \cite{Nguyen-Tuong2010}, the 
computational complexity of this approach scales quadratically with the number of training samples, 
such that exact inference typically becomes too slow for online learning in control loops. 

A common approach in robotics applications to overcome this issue of GPs bases on the divide and 
conquer principle of splitting up the data set and training multiple local GP models \cite{Nguyen-Tuong2009}.
Locally growing random trees of Gaussian processes (Log-GPs) are a recently proposed method following
this idea, which have been demonstrated to achieve update and prediction rates necessary for online
learning within control loops, while preserving many beneficial properties of exact GP inference 
\cite{Lederer2021b}. This method constructs a tree, whose leaf nodes contain locally active Gaussian
process models with a  maximum number $\bar{N}\in\mathbb{N}$ of training samples. Given a tree 
consisting only of the root node and no training samples, a binary tree is iteratively constructed 
with incoming training data $(\bm{x},y)$ using the following procedure:\looseness=-1
\begin{enumerate}
    \item Given a new training pair $(\bm{x},y)$, the tree is traversed until a leaf is reached 
    by going to the child of node $n$, 
    which is sampled from a Bernoulli distribution with probability~$p_n(\bm{x})$.
    \item If the leaf contains fewer than $\bar{N}$ training pairs continue with 5), else go to 3)
    \item add two child nodes to the current leaf node $n$ and distribute its training data by sampling from $p_n(\bm{x}^{(i)})$ for all $i=1,\ldots,\bar{N}$.
    \item Go to the child node sampled from~$p_n(\bm{x})$. 
    \item Add the training pair $(\bm{x},y)$ to the local GP model in the current leaf node and update its 
    hyperparameters.
\end{enumerate}
In this procedure, the Bernoulli distributions with probability functions $p_n(\cdot)$ define the regions
where local models are active. In order to quickly determine the probabilities $p_n(\bm{x})$, saturating 
linear functions of the form \pagebreak
\begin{align}
\label{eq:satlin}
p_n(\bm{x})\!=\!&\begin{cases}
0&\text{if } x_{j_n}< s_n-\frac{o_n}{2}\\
\frac{x_{j_n}\!-s_n}{o_n}\!+\!\frac{1}{2}& \text{if } s_n\!-\!\frac{o_n}{2}\!\leq\! x_{j_n}\!\leq\! s_n\!+\!\frac{o_n}{2}\\
1& \text{if } s_n+\frac{o_n}{2}< x_{j_n}
\end{cases}
\end{align}
have been proposed \cite{Lederer2021b}, where $j_n$ denotes the dimension in which the state space is split, 
$s_n$ denotes the value of the splitting plane, and $o_n$ corresponds to the size of the region where both 
local GPs are active to ensure a smooth global model. 
The splitting dimension $j_n$ can be chosen, e.g., as the dimension with the maximum spread in the training 
data, while a simple choice for the position of the splitting plane $s_n$ is the mean of the data in 
dimension $j_n$. Finally, the size of the overlapping region $o_n$ is typically chosen to be fixed
ratio of the extension the local models.

In order to retain a low computational complexity in the hyperparameter optimization in step 5), we merely 
perform a single gradient-based optimization step of the log-likelihood, whose partial derivatives are given by
\begin{equation}
	\frac{\partial}{\partial \theta_i}\!\log p(\bm{y}|\bm{X},\bm{\theta})\!=\! 
	\frac{1}{2}\left(\!\bm{y}^T\tilde{\bm{K}}\frac{\partial \bm{K}}{\partial \theta_i}\tilde{\bm{K}}\bm{y}\!-\!\mathrm{tr}\left(\tilde{\bm{K}}\frac{\partial \bm{K}}{\partial \theta_i}\!\right)\!\right)\!.\!
\end{equation}
Hence, hyperparameters are adapted online in step 5) using update rules of the form
\begin{align}\label{eq:rprop}
	\tilde{\bm{\theta}}^{n+1} &= \tilde{\bm{\theta}}^{n}+\bm{\psi}\left(\nabla_{\tilde{\bm{\theta}}}\log p(\bm{y}|\bm{X},\bm{\phi}(\tilde{\bm{\theta}}))\right)\Delta_t
\end{align}
where the step width $\Delta_t\in\mathbb{R}_+$ and the direction function $\bm{\psi}:\mathbb{R}^{\rho}\rightarrow\mathbb{R}^{\rho}$ can be defined to realize commonly 
used optimization schemes, 
e.g., steepest ascent or conjugate gradient \cite{Pedregal2004}. Since the computational complexity of the
update step \eqref{eq:rprop} only depends on the number of training samples in a local model but is 
independent of the overall amount of data, the update complexity of LoG-GPs remains logarithmic with this online
hyperparameter adaptation scheme.

Based on the constructed binary tree, the predictions of local GP models can be efficiently aggregated. 
For this purpose, given a test point $\bm{x}$, the probabilities $p_n(\bm{x})$ along a branch of the 
tree are multiplied to obtain the weight $w_m(\bm{x})$ of leaf node $m$. Then, standard aggregation schemes
such as mixtures of experts \cite{Tresp2001}
\begin{align}\label{eq:moe}
    \tilde{\mu}(\bm{x})=\sum\limits_{m=1}^M w_m(\bm{x})\mu_m(\bm{x})
\end{align}
where $M$ denotes the number of leaf nodes, 
can be employed to calculate the aggregated mean $\tilde{\mu}(\bm{x})$. Since the probabilities $p_n(\cdot)$ 
are usually chosen such that only few local models are active at the same test point, the tree structure 
can be again exploited for computing the weights $w_m(\bm{x})$, yielding a $\mathcal{O}(\log^2(N))$ complexity
for the computation of predictions under weak assumptions \cite{Lederer2021b}. Thereby, LoG-GPs can achieve 
the high prediction and update rates as required for learning in many control loops.

%%%%%%%%%%%%%%%%%%%%%%%%%%%%%%%%%%%%%%%%%%%%%%%%%%%%%%%%%%%%%%%%%%%%%%%%%%%%%%%%%%%%%%%%%%%%%%%%%%%%%%%%%%%%%%%%%%
%%%%%%%%%%%%%%%%%%%%%%%%%%%%%%%%%%%%%%%%%%%%%%%%%%%%%%%%%%%%%%%%%%%%%%%%%%%%%%%%%%%%%%%%%%%%%%%%%%%%%%%%%%%%%%%%%%
\section{INDIVIDUALIZED CONTROL USING GAUSSIAN PROCESS BASED ONLINE LEARNING}
\label{sec:theory}

In order to allow for an individualized treatment of patients, it is necessary to infer personalized 
models. Even though these models could be learned offline using data obtained during calibration phases 
conducted before the actual rehabilitation exercises, this is often perceived as cumbersome by 
patients. Therefore, we propose an online learning control law based on LoG-GPs which achieves
an adaptation to individual patients in virtually real-time. 

\begin{figure}
\centering
\begin{tikzpicture}[scale=0.79, every node/.style={scale=0.79}]

\fill[blue!20] (-6.5,-2.3) -- (2.3,-2.3) -- (2.3,0.75) -- (-2.2,0.75) -- (-2.2,1.5) -- (-5.7,1.5) -- (-5.7,3.47) -- (1.45,3.47) -- (1.45,0.75) -- (2.3,0.75) -- (2.3,4.0) -- (-6.5,4.0) -- (-6.5,-2.3);

\fill[red!20,draw=red!20] (1.45,0.75) -- (-2.2,0.75) -- (-2.2,1.5) -- (-5.7,1.5) -- (-5.7,3.47) -- (1.45,3.47) -- (1.45,0.75);

\node[align=center,draw=black, rectangle,minimum height=1.4cm, minimum width=2.1cm, rounded corners] at (0,0) {human-robot\\ system \eqref{eq:sys}};

\node[align=center,draw=black, rectangle,minimum height=1.4cm, minimum width=2.1cm, rounded corners] at (-4.2,-1.5) {CTC\\ policy\\ \eqref{eq:CTC}};

\node[align=center,draw=black, rectangle,minimum height=1.4cm, minimum width=2.1cm, rounded corners] at (-4.2,0) {PD\\ controller\\  \eqref{eq:PD}};

\node[align=center,draw=black, rectangle,minimum height=1.4cm, minimum width=2.1cm, rounded corners] at (0,1.5) {inverse\\ dynamics\\ \eqref{eq:inv_dyn}};

\node[align=center,draw=black, rectangle,minimum height=1.4cm, minimum width=2.1cm, rounded corners] at (-4.2,1.5) {LoG-GP\\ model\\  \eqref{eq:moe}};

\draw (-2.45,0) circle (0.15cm);
\fill (-1.75,0) circle (0.1cm);

\draw[>=latex, ->] (-3.15,0) -- (-2.6,0);
\draw[>=latex, ->] (-3.15,1.15) -- (-2.45,1.15) -- (-2.45,0.15);
\draw[>=latex, ->] (-3.15,-1.5) -- (-2.45,-1.5) -- (-2.45,-0.15);
\draw[>=latex, ->] (-2.3,0) -- (-1.05,0);

\draw[>=latex, ->] (1.05,0) -- (2.75,0);

\fill (1.9,0) circle (0.1cm);
\fill (1.9,1.5) circle (0.1cm);
\draw[>=latex, ->] (1.9,0) -- (1.9,1.5) -- (1.05,1.5);
\draw[>=latex, ->] (1.9,1.5) -- (1.9,2.95) -- (-2.05,2.95);
\draw[>=latex, ->]  (-2.85,2.95)-- (-4.45,2.95) -- (-4.45,2.2);

\draw (-1.75,1.5) circle (0.15cm);
\node at (-1.45,1.7) {$-$};
\draw[>=latex, ->] (-1.75,0) -- (-1.75,1.35);
\draw[>=latex, ->] (-1.05,1.5) -- (-1.6,1.5);
\draw[>=latex, ->] (-1.75,1.65) -- (-1.75,2.45) -- (-2.05,2.45);
\draw[>=latex, ->] (-2.85,2.45) -- (-3.95,2.45) -- (-3.95,2.2);

\draw (-2.85,2) rectangle (-2.05,3.4);
\draw (-2.85,2.7) -- (-2.65,2.7);
\draw (-2.05,2.7) -- (-2.25,2.7);
\draw (-2.25,2.7) -- (-2.55,2.9);
\draw[>=latex, ->] (-2.55,3.3) -- (-2.55,2.95);
\node at (-2.3,3.15) {$t_k$};

\fill (1.9,2.95) circle (0.1cm);
\draw[>=latex, ->] (1.9,2.95) -- (1.9,3.55) -- (-6.1,3.55) -- (-6.1,0.15);
\draw[>=latex, ->] (-5.95,0) -- (-5.25,0);
\fill (-6.1,1.15) circle (0.1cm);
\draw[>=latex, ->] (-6.1,1.15) -- (-5.25,1.15);

\draw (-6.1,0) circle (0.15cm);
\node at (-5.8,0.25) {$-$};
\fill (-6.1,-1.5) circle (0.1cm);
\draw[>=latex, ->] (-6.8,-1.5) -- (-5.25,-1.5);
\draw[>=latex, ->] (-6.1,-1.5) -- (-6.1,-0.15);

\node[align=left] at (-7.1,-1.5) {$\bm{q}_{\mathrm{ref}}$\\$\dot{\bm{q}}_{\mathrm{ref}}$\\$\ddot{\bm{q}}_{\mathrm{ref}}$};
\node[align=left] at (2.9,0) {$\bm{q}$\\$\dot{\bm{q}}$\\$\ddot{\bm{q}}$};
\node at (-1.4,0.2) {$\bm{u}$};
\node at (-1.4,1.2) {$\hat{\bm{u}}$};
\node at (-3.2,3.1) {$\bm{x}_k$};
\node at (-3.2,2.6) {$\bm{y}_k$};

\node[blue] at (-5.9,3.8) {control};
\node[red] at (-5.2,3.2) {learn};

\end{tikzpicture}
\vspace{-0.3cm}
\caption{Online learning control architecture with control components highlighted in blue and learning blocks illustrated in red. Noisy measurements of the system outputs and torque errors $\hat{\bm{u}}-\bm{u}$ are taken at times $t_k$ and used to infer a model online using LoG-GPs. Predictions of the LoG-GPs are employed as feedforward term in a CTC control law.}
\label{fig:overview}
\end{figure}
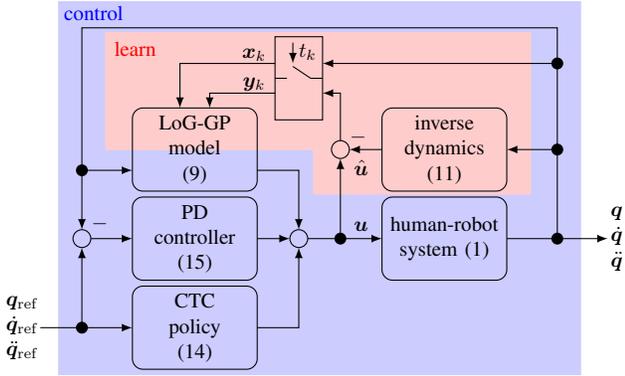

The proposed control architecture consisting of 
the online data generation and model inference as well as the feedforward and feedback control 
components is outlined in \cref{fig:overview}. Detailed information about the learning procedure can
be found in \cref{subsec:data}, while explanations regarding the individualized control law are 
provided in \cref{subsec:control}.

\subsection{Online Data Generation and Model Inference}\label{subsec:data}

In order to account for the individual human-induced dynamics in the controller, it is necessary to infer a model online
from data. In an idealized scenario, direct measurements of the torques generated by humans would be measured,
i.e., $\bm{y} = \bm{f}_i(\bm{x})$, 
such that a model of the patient's dynamics can be learned from the training samples $\bm{x}$, $\bm{y}$. However, 
due to the considered lack of force torque sensors for measuring the interaction forces as stated 
in \cref{ass:measure}, we cannot directly employ measurements of $\bm{f}_i(\cdot)$ for training GP models. 
This problem can straightforwardly be overcome by rearranging~\eqref{eq:sys}, 
such that
\begin{align}
    \bm{f}_i(\bm{x})=\bm{u}-\hat{\bm{u}}(\bm{x}),
\end{align}
where
\begin{align}\label{eq:inv_dyn}
    \hat{\bm{u}}(\bm{x})=\bm{H}(\bm{q})\ddot{\bm{q}}+\bm{C}(\bm{q},\dot{\bm{q}})\dot{\bm{q}}+\bm{g}(\bm{q})
\end{align}
denotes the inverse dynamics model. Since the applied torque $\bm{u}$ is determined by the employed control law and the parameters of the Euler-Lagrange system describing the robot dynamics are known as stated in \cref{ass:knowledgde}, $\hat{\bm{u}}(\bm{x})$ can directly be computed. Therefore, it remains to define a sampling rate $1/\tau$, $\tau\in \mathbb{R}_+$ at which measurements of $\bm{x}$ and $\bm{u}$ are taken, such that a training data set
\begin{align}
    \{(\bm{x}^{(k)}=\bm{x}(k\tau),\bm{y}^{(k)}=\bm{u}(k\tau)-\hat{\bm{u}}(\bm{x}(k\tau)))\}_{k=0}^K
\end{align}
for $K=\lfloor t/\tau \rfloor$ is aggregated based on the online measurements. Using these online data, we can update an independent GP for each target dimension of $\bm{y}^{(k)}$, i.e., for each $i=1,\ldots,d$ a LoG-GP is updated using a training pair $(\bm{x}^{(k)},y_i^{(k)})$ as explained in \cref{subsec:log-GP}. In order to employ these LoG-GPs in the control loop, their predictions are concatenated into the vector $\tilde{\bm{\mu}}(\bm{x})=[\mu_1(\bm{x})\ \cdots\ \mu_d(\bm{x})]^T$.

\begin{remark}
The proposed approach for learning a model of the individual patient dynamics in physical human-robot
interaction online is independent of a particular control law. Therefore, the choice of a particular control law does 
not affect the model inference approach.
\end{remark}

\subsection{Individualized Control Law}\label{subsec:control}

By exploiting the online learned model in the form of a feedforward control, it is straightforward 
to achieve a flexible adaptation to the individual dynamics of each patient without the need for 
additional calibration phases. Therefore, we propose the individualized control law
\begin{align}\label{eq:policy}
    \bm{u}=\bm{u}_{\mathrm{CTC}}(\bm{p})+\bm{u}_{PD}(\bm{e},\dot{\bm{e}})+\tilde{\bm{\mu}}(\bm{x}),
\end{align}
where the computed torque control 
\begin{align}\label{eq:CTC}
    \bm{u}_{\mathrm{CTC}}(\bm{p},\bm{p}_{\mathrm{ref}})=\bm{H}(\bm{q})\ddot{\bm{q}}_{\mathrm{ref}}+\bm{C}(\bm{q},\dot{\bm{q}})\dot{\bm{q}}_{\mathrm{ref}}+\bm{g}(\bm{q})
\end{align}
is used to compensate the nonlinear dynamics of the robotic system, and the PD controller
\begin{align}\label{eq:PD}
    \bm{u}_{PD}(\bm{e},\dot{\bm{e}})=-\bm{K}_p\bm{e}-\bm{K}_d\dot{\bm{e}}
\end{align}
with control gains $\bm{K}_p,\bm{K}_d\!\in\!\mathbb{R}^{d\times d}$
ensures the convergence of the tracking error $\bm{e}\!=\!\bm{q}\!-\!\bm{q}_{\mathrm{ref}}$ to a 
neighborhood of $\bm{0}$.\looseness=-1 

The PD gains allow a flexible trade-off between compliant behavior and high tracking accuracy,
with low gains $\bm{K}_p$, $\bm{K}_d$ leading to small control inputs of the PD controller~\eqref{eq:PD} 
in practice. Since the computed torque control \eqref{eq:CTC} is practically 
bounded under \cref{ass:ref} and the magnitude of LoG-GP predictions is restricted through the
observed torques \cite{Capone2021b}, which are applied by the patient, it is straightforward to see that the individualized control
law \eqref{eq:policy} with a sufficiently compliant PD controller can ensure safe interaction forces. 
This dependency of safety on the control gains is 
demonstrated experimentally in \cref{subsec:high-low-gain}.

\begin{remark}
Since the individualized control law \eqref{eq:policy} is based on a classical computed torque approach 
and uniform error bounds for predictions with LoG-GPs can be shown under weak assumptions \cite{Lederer2021b},
it is straightforward to analyze the stability of the closed-loop human-robot system using Lyapunov theory
analogously to, e.g., \cite{Helwa2019,Umlauft2020}. For reasons of brevity, a formal stability 
analysis is omitted here.
\end{remark}

\section{EXPERIMENTAL EVALUATION}\label{sec:experiment}
For the evaluation of the proposed learning-based controller, we perform experiments with a two DoF human-robot interaction setup, which is explained in \cref{subsec:setup}. First, the capacity of the method to successfully assist during a rehabilitation exercise whilst applying safe control outputs is shown in \cref{subsec:high-low-gain}. 
After this initial validation, the method is contrasted with a controller tuned for one individual, in \cref{subsec:tuned-GP}. Thereby, we demonstrate the learners ability to adapt to different human operators and provide personalized assistance. \looseness=-1

\subsection{Experiment Setup and Task Design}\label{subsec:setup}
\begin{figure}
\centering
\includegraphics[scale=0.112]{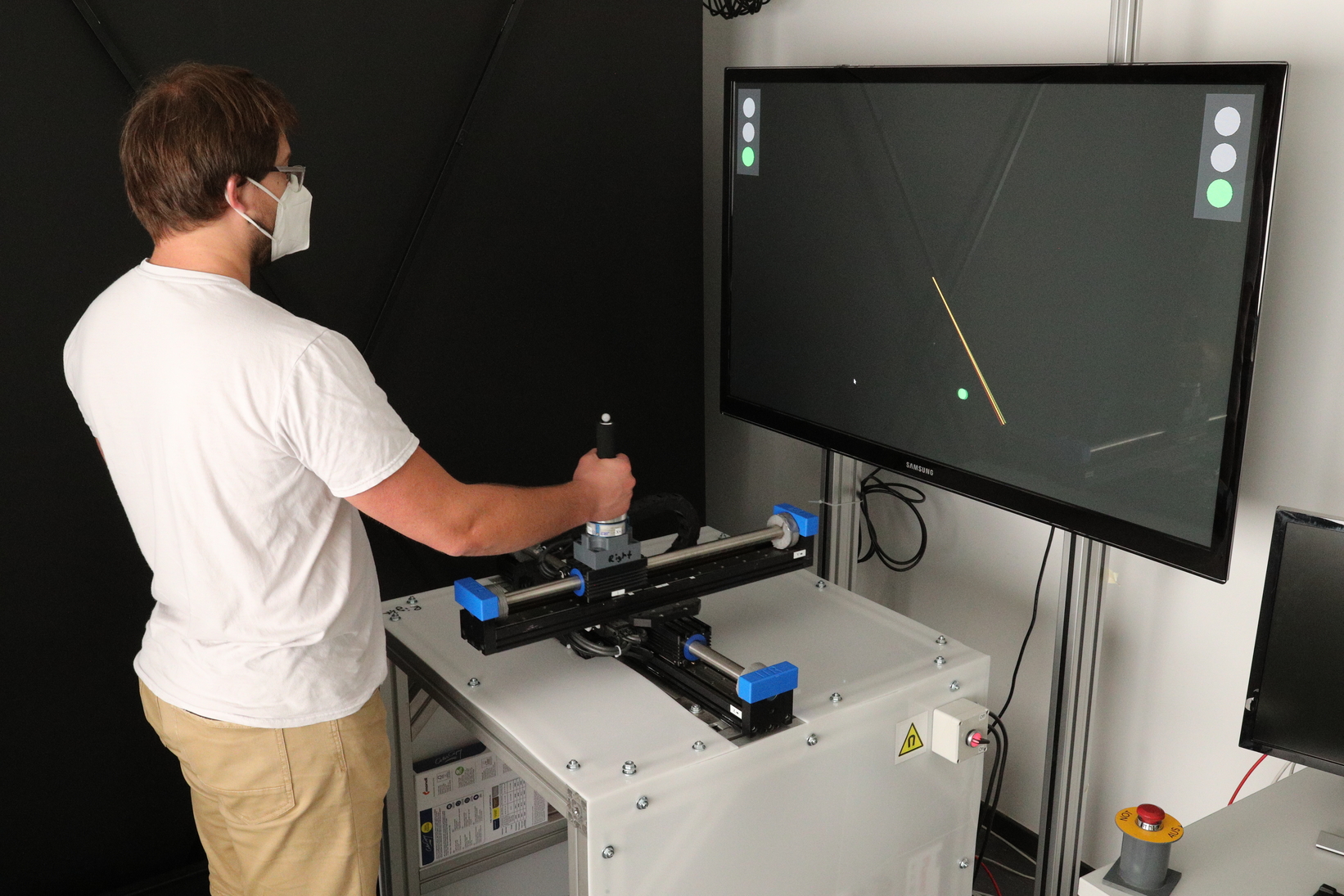}
\vspace{-0.3cm}
\caption{Reenactment of an individual performing the experiment task with the manipulandum. Consent for the publication of the image was obtained.\looseness=-1}
\label{fig:exp_setup}
\end{figure}
The experiments are executed on a two DoF manipulandum, which consists of two orthogonally mounted single rail stages (Copley Controls Thrustube Module), each driven by linear servo motors. Both rail stages are equipped with optical encoders that measure the position of a cart on the upper rail with \SI{1}{\micro\metre} precision. Additionally, a six DoF force-torque sensor (JR3-75M25) is mounted below the handle, through which the human interacts with the system, to measure forces in the horizontal plan. The force-torque sensor is strictly required to operate the device and is not used for the computation of control policies. Additionally, an inherent output force limitation is integrated as a safety measure, which guarantees that the interaction force applied to the human remains in safe regions, therefore, enabling experiments with aggressively tuned controllers. The device runs at \SI{4}{k\Hz} and the workspace of the complete apparatus spans \SI{\pm 0.20}{\metre} in both DoF. Visual feedback to perform the task is provided to the user through a screen placed behind the apparatus. The complete experiment setup and apparatus is shown in \cref{fig:exp_setup}. The task itself is designed as follows: \looseness=-1

\begin{figure}
\centering
\begin{tikzpicture}[scale=0.70, every node/.style={scale=0.70}]

\fill[black!80] (-4.65,-3.5)  rectangle (4.65,3.5);
\draw[black!10, rounded corners=15, ultra thick, dashed] (-3.0, -2.0) rectangle (3.0 , 2.0);

\draw[fill=green, align=center] (1.7,2.0) circle(0.25) node [xshift=-0.6cm, yshift=0.75cm, text=green]{\textbf{reference}\\ \textbf{position}};
\draw[orange, line width=1mm, align=center] (0,0) -- (4.45,3.1) node [midway, xshift=1.6cm, yshift=0.25cm, text=orange]{\textbf{visual}\\ \textbf{feedback}};

\draw[densely dashed, very thick, draw = white, opacity=0.9, fill=black!10, fill opacity=0.7, text opacity=1, align=center] (0.74,0.5) circle(0.25) node [xshift=0.6cm, yshift=-0.85cm, text=black!10]{\textbf{actual}\\ \textbf{position} \\ (not visible)};

\draw[very thick, draw = white] (0,0) circle(0.075) node []{\textbf{}};

\draw[black!10, very thick, align=center] (-2.75,-2.0) -- (-2.75,-2.3) node [xshift=0cm, yshift=-0.45cm, text=black!10]{\textbf{reference trajectory} \\ (not visible)};

\end{tikzpicture}
\vspace{-0.3cm}
\caption{Exemplary depiction of the task design. %of the experimental user study. 
The rounded rectangle drawn with the gray, dashed line represents the reference trajectory of the green circle, which the participant is instructed to track. Instead of the actual current position, depicted by the gray circle, the subject can only %see the polar angle shown by the orange line.
\looseness=-1 }
\label{fig:task}
\end{figure}
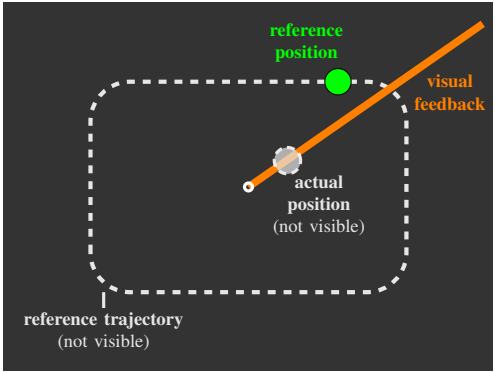

Standing in front of the apparatus and facing the screen, the subjects are instructed to track a green dot by moving the handle on top of the cart. In addition the participants are informed that different controllers will support them during task execution. Since the experiments are performed by healthy subjects, the provided visual feedback is artificially modified, therefore, limiting the participants' ability to successfully perform the task and thereby mimicking the physical limitations of an impaired patient. Specifically, the subjects do not see their current position in the task space entirely, but instead only the angle from the origin is visualized through a pointer. However, despite the limited feedback, their tracking performance is still evaluated on the position error in Cartesian coordinates. The task design and the visual feedback is depicted in \cref{fig:task}. Each run of the experiment begins at the same starting position for the green circle and consists of five repetitions of the reference trajectory. The complete experimental procedure can be split into two parts; first a training phase, followed by a test phase. During the initial training phase the participants get accustomed to the task and the assistance by performing one experiment run with each controller. Subsequently, the test phase begins, which consists of four experiment runs per controller. At every run a random controller variation is selected for assistance. If during any trial the workspace limit is reached, the device shuts down as a safety precaution and the run is evaluated as a failure. The failed runs are not repeated subsequently. \looseness=-1

\subsection{Compliance and Accuracy with Learning Control}\label{subsec:high-low-gain}
\pgfplotsset{
  /pgfplots/arrow/.style 2 args={
    legend image code/.code={
    \draw[|-|, thick] (0cm,0cm) -- (0.6cm,0cm);
      node[pos=0,#1]{}
      node[#2]{};%
    }
  }
}

\begin{table}[!t]
	\center
	\caption{PD control gains used in the experiments.}
		\begin{sc}
			\begin{tabular}{l c c c c}
				\toprule
				 & low gain& high-gain & GP & tuned gains\\
				\midrule
				$k_p$& $1$ & $600$ & $1$ & $35$\\
				$k_d$& $0.1$ & $60$ & $0.1$ & $3.5$\\
				\bottomrule
			\end{tabular}
		\end{sc}
	
	\label{tab:med_size}
\end{table}

In order to demonstrate the applicability of our approach, we conduct a user study with 9 healthy, right-handed participants between the age of 22 and 35. Participants signed a written informed consent, approved by the ethics committee of the medical faculty of the Technical University of Munich. During the experiment the operators perform a task whilst being assisted by three different variations of the proposed control architecture. A high-gain, low-gain and GP variation with PD control-parametrization according to \cref{tab:med_size} and diagonal gain matrices $\bm{K}_p=k_p\bm{I}$, $\bm{K}_d=k_d\bm{I}$, $k_p,k_d\in\mathbb{R}_+$. For the high-gain and low-gain controller there is no learning GP and they only differ with regards to the gains used for the PD controller \eqref{eq:PD}. Differently, the GP controller uses the individualized control law \eqref{eq:policy} including the proposed learning-based controller with small gains for the PD controller. The online hyperparameter adaptation \eqref{eq:rprop} is realized using RPROP \cite{Riedmiller1993} since it has been demonstrated to exhibit lower computational complexity and faster convergence compared to other gradient-based optimization schemes \cite{Blum2013}. All three controllers have the same CTC policy \eqref{eq:CTC} and the GP runs at a update and prediction rate of \SI{200}{\Hz}, resulting in approximately 10000 training samples at the end of one experimental run.\looseness=-1

\begin{figure}
\centering
\begin{tikzpicture} 
\begin{axis}[name=plot1,height=3.25cm, legend columns=2, legend entries={low-gain, high-gain, GP, standard deviation},	legend style={at={(0.68,0.975)},anchor=north,font=\scriptsize},  xtick={0,1,2}, xticklabel style={align=center, font=\footnotesize}, 
            xticklabels={,,}, 
            ymin=-0.0, ymax=125.0,
            ylabel style={align=center, font=\footnotesize},
            ylabel={{sum of abs. \\ error $[\SI{}{\metre}]$ }},
            every axis plot/.append style={
            ybar,
            bar width=.2,
            bar shift=0pt,
            fill
            }]
    
    	\addlegendimage{area legend, fill=blue!70!white, fill opacity=1}
		\addlegendimage{area legend, fill=red!80!white, fill opacity=1}
		\addlegendimage{area legend, fill=nicegreen!90!white, fill opacity=1}
		\addlegendimage{arrow} 
    
    % low gain bar plot
    \addplot+ [ only marks , no marks, mark options={scale=0.75, fill=blue!30!black, blue!30!black}, 
                blue!30!black, fill=blue!70!white, ybar, error bars/.cd, y dir=both, y explicit, 
                error bar style={line width=1pt, blue!30!black},
                error mark options={ rotate=90, blue!30!black, mark size=4pt, line width=1pt}] 
        coordinates {(0,66.1064) +- (-36.452,36.452)};
    % high gain bar plot
    \addplot+ [ only marks , no marks, mark options={scale=0.75, fill=red!30!black, red!30!black}, 
                red!30!black, fill=red!80!white, ybar, error bars/.cd, y dir=both, y explicit, 
                error bar style={line width=1pt, red!30!black},
                error mark options={ rotate=90, red!30!black, mark size=4pt, line width=1pt}]  coordinates {(1,6.56719) +- (-6.32288,6.32288)};
    % GP bar plot
    \addplot+ [ only marks , no marks, mark options={scale=0.75, fill=nicegreen!40!black, nicegreen!40!black}, 
                nicegreen!40!black, fill=nicegreen!90!white, ybar, error bars/.cd, y dir=both, y explicit, 
                error bar style={line width=1pt, nicegreen!40!black},
                error mark options={ rotate=90, nicegreen!40!black, mark size=4pt, line width=1pt}]  coordinates {(2,31.658) +- (-11.2727,11.2727)};
\end{axis}
\begin{axis}[at=(plot1.south), anchor=south,yshift=-1.8cm, height=3.25cm, 
            xtick={0,1,2}, xticklabel style={align=center, font=\footnotesize}, xticklabels = {{low-gain \\ controller},{high-gain \\ controller},{GP \\ controller}},
            ymin=-0.0, ymax=19.0, 
            ylabel style={align=center, font=\footnotesize},
            ylabel={{max. control \\ force $[\SI{}{\newton}]$ }},
            every axis plot/.append style={
            ybar,
            bar width=.2,
            bar shift=0pt,
            fill
            }]  
            
    \addplot+ [ only marks , no marks, mark options={scale=0.75, fill=blue!30!black, blue!30!black}, 
                blue!30!black, fill=blue!70!white, ybar, error bars/.cd, y dir=both, y explicit, 
                error bar style={line width=1pt, blue!30!black},
                error mark options={ rotate=90, blue!30!black, mark size=4pt, line width=1pt}]
        coordinates {(0,2.64894) +- (-0.02572,0.02572)};
    \addplot+ [ only marks ,no marks, mark options={scale=0.75, fill=red!30!black, red!30!black}, 
                red!30!black, fill=red!80!white, ybar, error bars/.cd, y dir=both, y explicit, 
                error bar style={line width=1pt, red!30!black},
                error mark options={ rotate=90, red!30!black, mark size=4pt, line width=1pt}] coordinates {(1,13.2476) +- (-5.2557,5.2557)};
    \addplot+ [ only marks ,no marks, mark options={scale=0.75, fill=nicegreen!40!black, nicegreen!40!black}, 
                nicegreen!40!black, fill=nicegreen!90!white, ybar, error bars/.cd, y dir=both, y explicit, 
                error bar style={line width=1pt, nicegreen!40!black},
                error mark options={ rotate=90, nicegreen!40!black, mark size=4pt, line width=1pt}] coordinates {(2,7.9059) +- (-4.2665,4.2665)};
\end{axis}
\end{tikzpicture}\vspace{-0.3cm}
\caption{Top: Mean and standard deviation of the summed absolute error for all participants over low-gain, high-gain and GP controller configuration. Bottom: Mean and standard deviation of the maximum applied control force norm at each run. The GP controller trades-off applied maximum control force and tracking performance, while the low-gain and high-gain controller either exhibit low tracking performance or overly large control forces.}
\label{fig:stat_error}
\end{figure}
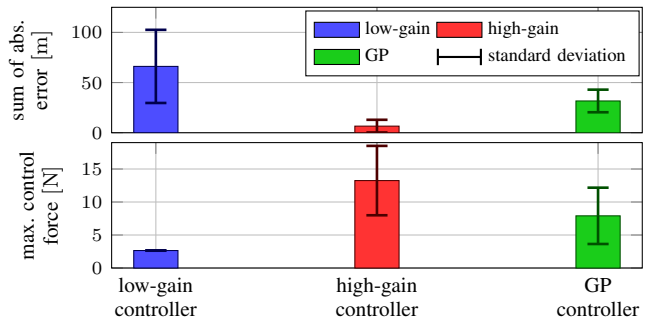

The analysis of the experimental results are shown in \cref{fig:stat_error}. The top bar plots depict the tracking performance by comparing the mean and standard deviation of the summed absolute error for each controller. 
It is clearly visible that the low-gain controller leads to the worst tracking accuracy and exhibits a notable variation between subjects, which is apparent due to the large standard deviation. In fact, four subjects reached the workspace limit at least once when they were assisted by the low-gain controller, resulting in a total of eight failed runs, which demonstrates the issues of highly compliant controllers in ensuring a successful task execution. Since the failed runs are particularly short they are not included in the analysis depicted in \cref{fig:stat_error}. While the high gain controller does not suffer from failed trials and exhibits the best control performance, it can result in uncomfortable interaction forces, which may ultimately become unsafe. This can be observed at the bottom of \cref{fig:stat_error} depicting the applied maximum forces, which are largest for the high gain configuration. In contrast to these PD control laws, the proposed online learning controller adapts itself to each participant. Since the labels used during training of the GP correspond to the human-generated torques, it is highly unlikely that higher control forces are generated than the human operator can manage. This is also demonstrated by the experiments as shown in \cref{fig:stat_error}. The applied maximum control forces of the GP controller remain significantly smaller compared to the high-gain configuration, while the tracking performance is strongly improved over the low-gain controller. Therefore, it is demonstrated that the proposed online learning control scheme is capable of successful task execution while being safer than high-gain controllers due to smaller interaction forces.\looseness=-1

\subsection{Personalized Assistance through Online Learning}\label{subsec:tuned-GP}

\pgfdeclarepatternformonly{northeast}{\pgfqpoint{-1pt}{-1pt}}{\pgfqpoint{4pt}{4pt}}{\pgfqpoint{3pt}{3pt}}%
{
  \pgfsetlinewidth{0.3pt}
  \pgfpathmoveto{\pgfqpoint{0pt}{0pt}}
  \pgfpathlineto{\pgfqpoint{3.1pt}{3.1pt}}
  \pgfusepath{stroke}
}

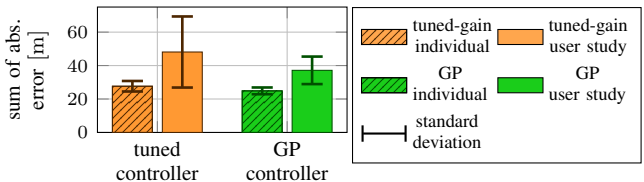
\begin{figure}
\centering
\begin{tikzpicture} 
\begin{axis}[name=plot1,height=3.25cm, width = 4.9cm, legend columns=2, 
            legend style={align=center, at={(1.6,1)},anchor=north,font=\scriptsize},
            legend entries={{tuned-gain \\ individual}, {tuned-gain \\ user study}, {GP \\ individual}, {GP \\ user study}, {standard \\ deviation}},
            xtick={-0.075,0.575,2}, xticklabel style={align=center, font=\footnotesize}, 
            xticklabels={,,}, xticklabel style={align=center, font=\footnotesize}, xticklabels = {{tuned \\ controller},{GP \\ controller}},
            xmin=-0.375, xmax=0.875, 
            ymin=-0.0, ymax=75.0,
            ylabel style={align=center, font=\footnotesize},
            ylabel={{sum of abs. \\ error $[\SI{}{\metre}]$ }},
            every axis plot/.append style={
            ybar,
            bar width=.2,
            bar shift=0pt,
            fill
            }]
        \addlegendimage{area legend, fill=orange!70!white, postaction={pattern=northeast, pattern color=orange!30!black}, fill opacity=1}
		\addlegendimage{area legend, fill=orange!70!white, fill opacity=1}
		\addlegendimage{area legend, fill=nicegreen!90!white, postaction={pattern=northeast, pattern color=nicegreen!40!black}, fill opacity=1}
		\addlegendimage{area legend, fill=nicegreen!90!white, fill opacity=1}
		\addlegendimage{arrow}   

    \addplot+ [ only marks ,no marks, mark options={scale=0.75, fill=orange!30!black, orange!30!black}, 
                orange!30!black, fill=orange!80!white, 
                postaction={pattern=northeast, pattern color=orange!30!black}, 
                ybar, error bars/.cd, y dir=both, y explicit, 
                error bar style={line width=1pt, orange!30!black},
                error mark options={ rotate=90, orange!30!black, mark size=4pt, line width=1pt},    
                ] coordinates {(-0.2,27.6342) +- (-3.1466,3.1466)};
    \addplot+ [ only marks ,no marks, mark options={scale=0.75, fill=orange!30!black, orange!30!black}, 
                orange!30!black, fill=orange!80!white, ybar, error bars/.cd, y dir=both, y explicit, 
                error bar style={line width=1pt, orange!30!black},
                error mark options={ rotate=90, orange!30!black, mark size=4pt, line width=1pt}] coordinates {(0.05,48.1107) +- (-21.2561,21.2561)};
                
    \addplot+ [ only marks ,no marks, mark options={scale=0.75, fill=nicegreen!40!black, nicegreen!40!black}, 
                nicegreen!40!black, fill=nicegreen!90!white, 
                postaction={pattern=northeast, pattern color=nicegreen!40!black}, 
                ybar, error bars/.cd, y dir=both, y explicit, 
                error bar style={line width=1pt, nicegreen!40!black},
                error mark options={ rotate=90, nicegreen!40!black, mark size=4pt, line width=1pt}] coordinates {(0.45,24.8749) +- (-2.0195,2.0195)};
    \addplot+ [ only marks ,no marks, mark options={scale=0.75, fill=nicegreen!40!black, nicegreen!40!black}, 
                nicegreen!40!black, fill=nicegreen!90!white, ybar, error bars/.cd, y dir=both, y explicit, 
                error bar style={line width=1pt, nicegreen!40!black},
                error mark options={ rotate=90, nicegreen!40!black, mark size=4pt, line width=1pt}] coordinates {(0.7,37.1473) +- (-8.2305,8.2305)};
\end{axis}
\end{tikzpicture}\vspace{-0.3cm}
\caption{Mean and standard deviation of the summed absolute error for one surrogate participant and the complete user study. 
While the tuned PD controller and GP controller perform comparably for the surrogate individual, only the GP controller adapts in the user study to personalize the assistance required for acceptable tracking performance.}
\label{fig:tuned_error}
\end{figure}

In order to demonstrate the benefits of the proposed learning controller, we compare it to a PD controller with tuned gains. Due to the lack of a better procedure for adaptation in human-robot interaction, the gains of the PD controller are tuned heuristically to balance the applied forces and the resulting tracking performance. As practical considerations prevent a tuning with all participants, one individual is chosen instead. However, since the PD controller needs to be safe for all users, a cautious tuning is preferred, which tends to result in lower control gains. The best trade-off is obtained for the parametrization depicted on the right side of \cref{tab:med_size}.
\looseness=-1
 
\begin{figure}
\centering
\begin{tikzpicture} 
\begin{axis}[name=plot1,height=3.25cm, width = 4.25cm, legend columns=2, 
            legend style={align=center, at={(1.06,1.35)},anchor=north,font=\scriptsize},
            legend entries={tuned-gain user study\quad, GP user study},
            xtick={-0.125,0.125}, xticklabel style={align=center, font=\footnotesize}, 
            xticklabels={PD,GP}, 
            xmin=-0.375, xmax=0.375, 
            ymin=-0.0, ymax=22.5,
            ylabel style={align=center, font=\footnotesize},
            ylabel={{intra-subject \\ variation $[\SI{}{\metre}]$ }},
            every axis plot/.append style={
            ybar,
            bar width=.2,
            bar shift=0pt,
            fill
            }]
		\addlegendimage{area legend, fill=orange!70!white, fill opacity=1}
		\addlegendimage{area legend, fill=nicegreen!90!white, fill opacity=1}   

    \addplot+ [ only marks ,no marks, mark options={scale=0.75, fill=orange!30!black, orange!30!black}, 
                orange!30!black, fill=orange!80!white,
                ybar, error bars/.cd, y dir=both, y explicit, 
                error bar style={line width=1pt, orange!30!black},
                error mark options={ rotate=90, orange!30!black, mark size=4pt, line width=1pt},    
                ] coordinates {(-0.125,6.431)};% +- (-3.1466,3.1466)};
    \addplot+ [ only marks ,no marks, mark options={scale=0.75, fill=nicegreen!40!black, nicegreen!40!black}, 
                nicegreen!40!black, fill=nicegreen!90!white, ybar, error bars/.cd, y dir=both, y explicit, 
                error bar style={line width=1pt, nicegreen!40!black},
                error mark options={ rotate=90, nicegreen!40!black, mark size=4pt, line width=1pt}] coordinates {(0.125,3.547)}; % +- (-21.2561,21.2561)};
\end{axis}
\begin{axis}[at=(plot1.east), anchor=east,xshift=3.0cm,height=3.25cm, width = 4.25cm,	 
            xtick={-0.125,0.125}, xticklabel style={align=center, font=\footnotesize}, xticklabels = {PD,GP},
            xmin=-0.375, xmax=0.375,
            ymin=-0.0, ymax=22.5,
            ylabel near ticks, yticklabel pos=right,
            ylabel style={align=center, font=\footnotesize},
            ylabel={{inter-subject\\ variation $[\SI{}{\metre}]$ }},
            every axis plot/.append style={
            ybar,
            bar width=.2,
            bar shift=0pt,
            fill
            }]

    \addplot+ [ only marks ,no marks, mark options={scale=0.75, fill=orange!30!black, orange!30!black}, 
                orange!30!black, fill=orange!80!white, 
                ybar, error bars/.cd, y dir=both, y explicit, 
                error bar style={line width=1pt, orange!30!black},
                error mark options={ rotate=90, orange!30!black, mark size=4pt, line width=1pt}] coordinates {(-0.125,21.3555)};% +- (-0.0828,0.0828)};
    \addplot+ [ only marks ,no marks, mark options={scale=0.75, fill=nicegreen!40!black, nicegreen!40!black}, 
                nicegreen!40!black, fill=nicegreen!90!white, ybar, error bars/.cd, y dir=both, y explicit, 
                error bar style={line width=1pt, nicegreen!40!black},
                error mark options={ rotate=90, nicegreen!40!black, mark size=4pt, line width=1pt}] coordinates {(0.1275,7.78)};% +- (-0.6575,0.6575)};
        
\end{axis}
\end{tikzpicture}\vspace{-0.3cm}
\caption{Left: Average of intra-subject standard deviation in tracking error. Right: Inter-subject standard deviation of participant specific average tracking error. The adaptability of the GP controller is reflected in the smaller intra- and inter-subject variation.}
\label{fig:intra_inter}
\end{figure}
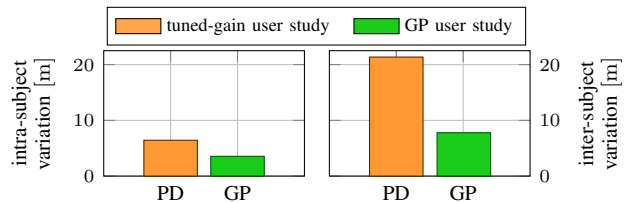

The tuned PD controller and the proposed online learning control law are evaluated as described in \cref{subsec:high-low-gain}, which leads to the results depicted in \cref{fig:tuned_error}. For the individual, the two controllers perform comparably well with regards to tracking performance, with the GP controller leading to slightly better tracking. However, the observed difference in tracking error is insignificant, since it lies within the statistical variation, and can be attributed to the cautious tuning of the PD controller. When deploying the tuned PD controller to previously unobserved individuals and comparing the performance to the learning-based GP controller in a user study, it can be seen that the tuned PD controller performs significantly worse.
Similarly, the GP controller on average results in higher tracking errors in the user study than for the surrogate individual. However, the increase in mean tracking error is larger for the tuned PD controller with a substantial growth in the standard deviation. Therefore, the tuned PD controller leads to inconsistent tracking results, which can be attributed to the different levels of task proficiency of the participants. 
This becomes even clearer when looking 
at the intra- and inter-subject variation of the tracking error as depicted in \cref{fig:intra_inter}. While the 
participants exhibit a similar variation of the tracking error among the experiment runs for both controllers, the variation of the tracking error between different subjects is significantly larger for the tuned PD controller. This is due to the inability of the PD control law to adapt to unproficient participants, which require more guidance to execute the task properly. In contrast, the GP controller provides personalized support for each subject, such that the variation between different participants can be reduced.\looseness=-1

\section{Conclusion}\label{eq:conclusion}
This paper introduces a novel online learning control method for personalized rehabilitation robotics. %To the best of the authors knowledge, 
This is the first time that GPs are used directly in the control loop during physical human-robot collaboration. Furthermore, the proposed method facilitates personalized assistance via online learning with previously unseen update rates. The approach is validated in an experimental user study, which confirms the adaptation capabilities of the learning-based controller.
%%%%%%%%%%%%%%%%%%%%%%%%%%%%%%%%%%%%%%%%%%%%%%%%%%%%%%%%%%%%%%%%%%%%%%%%%%%%%%%%
\section{ACKNOWLEDGMENTS}

This work has received funding from the Horizon 2020 research and innovation programme of the European Union under grant agreement n$^\circ$ 871767 of the project ReHyb: Rehabilitation based on hybrid neuroprosthesis. A. L. gratefully  acknowledges  financial  support from  the German Academic Scholarship Foundation.

\bibliographystyle{IEEEtran}
\bibliography{IEEEabrv,myBib}             % bib file to produce the bibliography

\end{document}